\pdfoutput=1

\documentclass[11pt]{article}

\usepackage{configs/acl}

\usepackage{geometry}
\usepackage[linesnumbered,ruled,vlined]{algorithm2e}
\usepackage{multirow}
\usepackage{tabularx}
\usepackage{makecell}
\usepackage{amssymb} 
\usepackage{enumitem}
\usepackage{tcolorbox}
\usepackage{amsmath}
\usepackage{amsfonts}
\usepackage[misc]{ifsym}

\tcbuselibrary{most}
\usepackage{times}
\usepackage{latexsym}

\usepackage[T1]{fontenc}

\usepackage[utf8]{inputenc}

\usepackage{microtype}

\usepackage{inconsolata}

\usepackage{graphicx}
\graphicspath{figures/}
\usepackage{amsmath,amsfonts,bm}
\usepackage{enumitem}
\usepackage{subfig}
\usepackage{wrapfig}
\usepackage{multirow}
\usepackage{multicol}
\usepackage{makecell}
\usepackage{extarrows}
\usepackage{xspace}
\usepackage[capitalize]{cleveref}
\usepackage{caption}
\usepackage[linesnumbered,ruled,vlined]{algorithm2e}
\usepackage{algpseudocode}
\usepackage{bbm}
\usepackage{float}
\usepackage{xcolor,colortbl}
\usepackage{diagbox}
\usepackage{verbatim}
\usepackage{framed}
\usepackage{tcolorbox}
\usepackage{booktabs}
\usepackage{titletoc}
\usepackage{appendix}
\usepackage{pifont}
\usepackage{acronym}
\usepackage{scalerel}

\def\titlelogo{\scaleobj{0.15}{\includegraphics{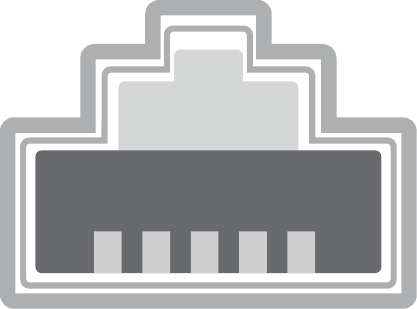}}}

\makeatletter
\DeclareRobustCommand\onedot{\futurelet\@let@token\@onedot}
\def\@onedot{\ifx\@let@token.\else.\null\fi\xspace}
\def\eg{\emph{e.g}\onedot} 

\def\ie{\emph{i.e}\onedot}

\def\etc{\emph{etc}\onedot} 

\def\wrt{w.r.t\onedot}

\makeatother

\newcommand{\one}{\ding{202}\xspace}
\newcommand{\two}{\ding{203}\xspace}
\newcommand{\three}{\ding{204}\xspace}

\makeatletter
\renewcommand{\paragraph}{%
  \@startsection{paragraph}{4}%
  {\z@}{0ex \@plus 0ex \@minus 0ex}{-1em}%
  {\normalfont\normalsize\bfseries}%
}
\makeatother

\acrodef{nlp}[NLP]{natural language processing}
\acrodef{vqa}[VQA]{Visual Question Answering}
\acrodef{cs}[CS]{coherence scoring}
\acrodef{avsd}[AVSD]{coherence scoring}
\acrodef{roi}[RoI]{Region-of-Interest}
\acrodef{llm}[LLM]{large language model}
\acrodef{tom}[ToM]{theory-of-mind}
\acrodef{bce}[BCE]{Binary Cross Entropy}
\acrodef{mse}[MSE]{mean squared error}
\acrodef{rag}[RAG]{retrieval augmented generation}

\newcommand{\model}{RAM\xspace}
\newcommand{\frsh}{FreshQA\xspace}
\newcommand{\mq}{MQuAKE\xspace}
\newcommand{\prompt}[1]{{\fontfamily{lmtt}\selectfont #1}}

\DeclareMathOperator*{\argmax}{arg\,max}

\crefname{figure}{Fig.}{Figs.}
\crefname{table}{Tab.}{Tabs.}
\crefname{section}{Sec.}{Secs.}

\title{\begin{tabular}{cl}
\setlength{\tabcolsep}{.5pt}
    \multirow{2}{*}{\protect\titlelogo{}} & RAM: Towards an Ever-Improving Memory System \\
    & by Learning from Communications
\end{tabular}} 

\author{Jiaqi Li, Xiaobo Wang, Wentao Ding, Zihao Wang, Yipeng Kang, Zixia Jia, Zilong Zheng\textsuperscript{\Letter} \\
  Beijing Institute for General Artificial Intelligence (BIGAI), Beijing, China \\
  \texttt{\{lijiaqi, wangxiaobo, dingwentao, wangzihao, kangyipeng, jiazixia, zlzheng\}@bigai.ai} \\}
  
\begin{document}
\maketitle

\begin{abstract}

We introduce an innovative RAG-based framework with an ever-improving memory. Inspired by humans' pedagogical process, \model utilizes recursively reasoning-based retrieval and experience reflections to continually update the memory and learn from users' communicative feedback, namely communicative learning. 
 Extensive experiments with both simulated and real users demonstrate significant improvements over traditional RAG and self-knowledge methods, particularly excelling in handling false premise and multi-hop questions. 
 Furthermore, \model exhibits promising adaptability to various feedback and retrieval methods, showcasing its potential for advancing AI capabilities in dynamic knowledge acquisition and lifelong learning.
\end{abstract}

\section{Introduction}

\begin{quote}
``Learning, flexibility, and attention to the partner are
obviously fundamental characteristics of the human way
of \emph{communicating}, and things simply could not move in
the human direction until they were present.''
\par\nobreak\hfill---\citet{tomasello2010origins} 
\end{quote}

Human learning, extended as a lifelong process, typically operates in a communicative and cooperative framework among people via different forms of interactions within the physical and social world, as evidenced by the above quotes of \citet{tomasello2010origins}. From toddlers to academic graduates, the learning process (referred to as pedagogy; \citet{shulman1987knowledge}) often involves two communicative roles: a student that perceives, reasons and learns over the communicative messages (\eg, conversations) to update his/her internal knowledge (\textit{a.k.a.} belief); and a teacher that delivers messages and provides feedback based on his/her professionalism. Such a learning paradigm, named communicative learning~(CL; \citet{yuan2023communicative}) or interactive learning~(IL; \citet{de1992interactive}), is considered natural and transparent along with numerous advantages that today's artificial intelligence (AI) systems seek to obtain, \eg, explainability~\cite{lakkaraju2022rethinking}, life-long grown skills~\cite{dalvi2022towards}, \etc. 

\begin{figure*}[t!]
    \centering
    \includegraphics[width=0.9\linewidth]{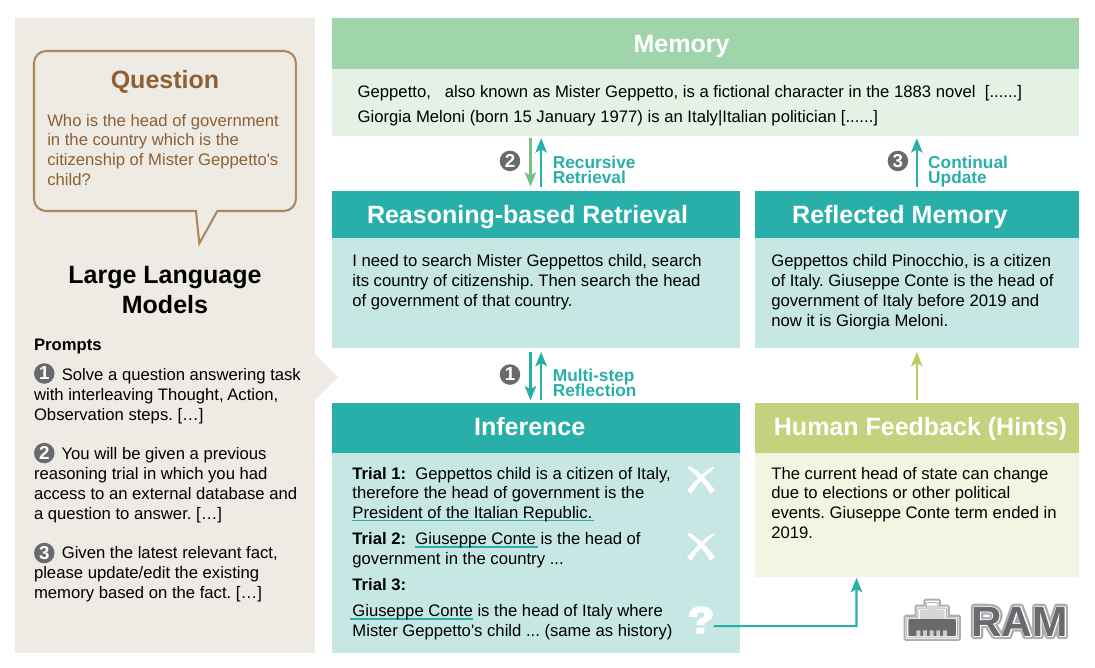}
    \caption{\textbf{Learning framework of \model}. Blue boxes indicate LLMs' in-context reasoning and the green box indicates feedback from external users. \one Given a new question, LLMs take multi-step reasoning and inference through self-reflection. If the current inference is the same as in previous trials, the human will provide additional hints as feedback to help LLMs better answer. \two Relevant knowledge is recursively retrieved from memory based on LLMs' reasoning. \three LLMs generate a reflected memory learning from the feedback and the ground truth to update the memory. All prompts are shortened for simplicity; refer to Appendix~\ref{appendix:prompts} for complete templates.}
    \label{fig:framework}
\end{figure*}

Empowered by the recent surge of large language models~(LLMs), many research works have attempted to build complicated AI agents to perform a spectrum of tasks with emergent capacities, such as Human-AI conversations, in-context reasoning~\cite{wei2022chain}, situated planning~\cite{wang2023describe, wang2023jarvis1}, \etc. Despite the compelling success, the \textit{immutability} and \textit{uninterpretability} of pre-trained language models yield non-trivial challenges in building communicatively-learned AI (CLAI) agents: (i) the \textbf{implicit knowledge representation} makes it hard to revise or edit pre-trained memory and may result in ``hallucinations''~\cite{lewis2020RAG}; (ii) the \textbf{limited context window size} constrains the potential to take in the entire context history as memory~\cite{li2023loogle}.
More recently, retrieval-augmented generation (RAG; \citet{lewis2020RAG}) is proposed to enable accessing and precisely manipulating the memory with a disentangled knowledge storage system; refer to $\S$\ref{sec:related_work} for details. However, conventional RAG augments LLMs with a \textit{static} and \textit{exterior} knowledge to address knowledge-intensive tasks. Fundamentally, the main challenge of building CLAI agents lies in determining when and how to update \textit{dynamic} and \textit{internal} knowledge given communicative feedback.

In this work, we introduce an innovative ever-improving memory system \textbf{\model}~(namely \textbf{\underline{R}}etrieval  \textbf{\underline{A}}ugmented  \textbf{\underline{M}}emory), in analogy to the fast-updated memory cache in computer systems (see \cref{fig:framework}). 
Without periodically re-training with a huge number of parameters, \model enables LLMs to obtain fresh knowledge and historical experience by dynamically improving and growing a continually updated memory through human communications ($\S$\ref{sec:framework}). 

Specifically, \model is composed of a recursive reasoning-based retrieval and a memory-reflection module: the former enables a faithful and self-refined reasoning trajectory throughout a recursively retrieval-based reasoning process (R$^3$; $\S$\ref{sec:r3}); the latter enriches the dynamic memory with current observations and user feedback for further self-improvement. To be close to real-world human teaching, we investigate different forms of common human feedback ($\S$\ref{sec:feedback}) to ablate \model's performance.
One close work to us is TeachMe~\cite{dalvi2022towards}, which aims to create a teachable QA system. However, 
TeachMe is restricted by solely adding user-supplied corrections to erroneous model beliefs for further retrieval, in which memory is far from well-updated and continually maintained.

In $\S$\ref{sec:exp}, comprehensive experiments with both simulated and real users empirically show that \model largely promotes the performance on various baselines, backbones and categories of knowledge. Specifically, under the evaluation metric of GPT4\_score on two datasets, \model achieves an average improvement of \textbf{30\%} over self-knowledge and \textbf{40\%} over RAG-only. Notably, \model exceeds methods with ground-truth updated memory. Moreover, \model excelled in false premise and multi-hop questions which current LLMs still struggle with. 
It is worth noting that for novel questions, the model gains \textbf{10-20\%} with self-reflected memory, allowing learning ever-changing knowledge and improving the reasoning capabilities in the long run. 
Finally, we make robust ablation studies to demonstrate the \textbf{generalization of \model} to different ways of teaching, feedback and retrieval chain types in practice.

\section{The \model Framework}\label{sec:framework}
In this section, we provide detailed descriptions of each key component of \textbf{\model} and how they collaborate in a unified framework. \cref{fig:framework} depicts the \model architecture. Detailed implementations are introduced in \cref{appendix:implement}.

\subsection{R$^3$: \underline{R}ecursive \underline{R}easoning-based \underline{R}etrieval}\label{sec:r3}

    

\begin{algorithm}[t] 
\small
    \caption{Reasoning process of \textbf{R$^3$}}\label{alg:memory}
    \KwIn{Query $Q$, memory buffer $\mathbf{M}$, ground truth $G$, action $A=\emptyset$, prompts \one and \two as in \cref{fig:framework}.}

    \For{$tr_i = 1$ \KwTo $N$}{
        \While{$A \neq \text{Finish}$}{
            \tcc{1. Reasoning Step}
            Get action $A$ and keyword $r$ with: $A, r \leftarrow \text{LLM}(Q; {\rm prompt}$\one$)$ \\
            \tcc{2. Retrieval Step}
            Get semantically relevant memory from $\mathbf{M}$: $m^* \leftarrow \arg\max_{m \in \mathbf{M}} \text{sim}(r, m)$ \\
             \tcc{3. Inference Step}
            Get inference result: $Inf \leftarrow \text{LLM}(m^*, \mathbf{FB}, Q; {\rm prompt}$\two$)$  \\
            \If{$Inf$ in historical inferences }{
                Get feedback with hints:
                $FB \leftarrow \text{Feedback}(Q, G)$  \\
            }
        }
        
        \If{$\text{sim}(Inf, G) > accept\_threshold$}{
            \textbf{break}
        }
    }
\Return $(m^*_i, Inf_i, FB_i), i \in [1,...,N]$
\end{algorithm}

Many previous methods have attempted to interleave reasoning and acting~\cite{yao2023react, shinn2023reflexion} in a chain-of-thought~(CoT) paradigm. More recently, such strategies have also been integrated with retrieval-based tools during the reasoning trace~\cite{luo2024rog}. However, these prompting methods, simply taking the query or its variants for RAG, failed to consider the \textit{dynamic semantics} during the reasoning process. For instance in \cref{fig:framework},  the initial key semantics of the query is ``Mister Geppeto\'s child''. Due to the nature of semantic matching, most likely the RAG engine will produce information \wrt ``Mister Geppeto'' or his child. However, by deeply going further along the reasoning trace, more extra information (\eg, ``head of Italy'') has to be taken into consideration. To this end, we propose \textbf{R}$^3$, a recursive CoT paradigm that prompts the model to iteratively retrieve and reason step-by-step to solve the question with a vector-based memory.

Given a query $Q$ and a memory buffer $M$ initialized with $K$ outdated knowledge $M=\{m_1, \cdots, m_K\}$. 
The whole R$^3$ process runs in a trial loop $\{tr_i, 1\leq i \leq N\}$.  At $tr_i$,  R$^3$ runs a sequential steps of Reason-Retrieval-Inference.
\textbf{Reason:} R$^3$ reasons on $Q$ and decomposes it into a few plausible actions $A$, \eg, ``\prompt{I need to search ... Then search ... }''. We then formalize the reasoning results to a sequence of actions \prompt{Search} using self-reflection. \textbf{Retrieval:} The action \prompt{Search} with a reflected keyword or phrase $r$ (\eg, ``\prompt{\underline{Minster Geppetos child}}'') is the result of reasoning on $Q$. R$^3$ retrieves the most relevant memory \wrt $m^* = \argmax_{m \in M} sim(r, m)$.  R$^3$ continues the Reason-Retrieval process until it finishes retrieval with an inference result. 
\textbf{Inference:} The model inferences on all the retrieved memory to obtain an inference result $Inf$ (\eg, ``\prompt{\underline{President of the Italian Republic}}'' in Trial. If $Inf$ is judged wrong, a new trial $tr_{i+1}$ for R$^3$ starts for more attempts. Otherwise, the model starts to update $M$ ($\S$\ref{sec:memory}). Inferences and feedback in all the trials are stored in the scratchpad for memory update. The overall process can be seen in Algorithm \ref{alg:memory}.


It has been shown from previous works~\cite{gao2024retrievalaugmented, yan2024corrective} that retrieval based on text similarity is far from enough to cope with complex tasks. Instead of a single call to retrieve the answer directly, R$^3$ proposed in  \model produces a promising paradigm to discover a \textbf{faithful reasoning trace} leading to a probable correct answer through multiple rounds of inference, reflection and interaction with external user/environment. Iterative retrieval provides \textbf{sufficient contexts and situated evidence} for forward reasoning, especially when coupled with restricted knowledge. In the meantime, interweaving reasoning on the track of retrieval helps to clarify the search direction, decompose the complex multi-retrieval task through planning, and narrow down the retrieval objective to get the final answer.

\textbf{Ask for help}\quad{} In \model, we maintain a list of historical inferences for each trial, \ie, $Inf = \{Inf_1, \cdots, Inf_N\}$. During R$^3$, if the current $Inf_i$ is semantically the same as any historical inference result, an ``ask-for-help'' mechanism is activated by querying human users for more hints to assist the reasoning process. It is possibly due to a lack of knowledge or a confined mindset on retrieval of the model which needs external help. In $\S$\ref{sec:feedback}, we propose different categories of human feedback in RAM.

\subsection{An Ever-Improving Memory}\label{sec:memory}
Existing RAG-based methods~\cite{tandon2022learning, madaan2023memoryassisted, sarch2023openended} suppose the knowledge is up-to-date~\cite{du2023static, zhong2023mquake} or can be acquired directly from a search engine~\cite{tu2023freshqa}. A naive solution is to add all feedback $FB$ to $M$ without dealing with knowledge fusion and alignment. The infinitely enlarging memory, however, makes the retrieval process time-consuming and inaccurate and is therefore infeasible for real-world and complicated contexts.

Let $M^{old}$ denote the initial memory buffer and $M^{cur}$ denote the current memory buffer, the memory update process in \model goes as follows. Given a new query $Q$, after the R$^3$ process ends with a correct $Inf$ or reaches $tr_N$, we localize the most relevant $m^* \in M^{cur}$ and locally edit it. Specifically, we start by collecting inferences and feedback in all the trials as context and prompting the LLM with the ground truth $G$ to generate a reflected memory $m^{R} = {\rm reflect}(G, Inf_1, FB_1, Inf_2, \cdots)$, where ${\rm reflect}(\cdot)$ denotes step \three in \cref{fig:framework}. Then we utilize the semantic similarity to localize the most relevant memory piece $m^* = \argmax_{m \in M^{cur}} sim(m, m^R) $ and update the memory buffer to $M^{upd}$ by replacing $m^*$ as $m^R$. 
The updated memory $M^{upd}$ is extensive and adaptive to ever-changing knowledge in the real world as the latest information is absorbed and outdated data is modified or discarded.

\subsection{Knowledge From Human Feedback}\label{sec:feedback}

Interactively learning from feedback is crucial for agents to avoid repeated errors from historical trials and accelerate the learning process with limited knowledge and capabilities. Closer to how humans learn, there are various types of feedback, that can benefit the model to get a reward/signal for its current performance, an explanation of past behaviors, instructions for future behavior learning through iterative interactions. We describe three different categories of human feedback in \model as below.

\paragraph{Feedback without explanation}
It serves as an automatic evaluator for $Inf_i$ in $tr_i$ for further retrieval and self-reflection.
We compute the semantic similarity between embeddings of $Inf$ with $G$ based on the pre-defined threshold. Using automatic similarity as $FB$ provides more flexibility than traditional n-gram matching while remaining comparatively accurate with lower costs than employing LLM itself as an evaluator.

\paragraph{Feedback with hints}

Instead of offering ground truth directly (or statements semantically the same), it is expected to conditionally provide either additional knowledge or a new direction for better retrieval based on previous scratchpads and given ground truth. It not only allows the model to learn the association among multiple relevant pieces of knowledge in a single problem but also teaches the model the way of thinking to continually promote its intricate reasoning capability from historical trials.
 


\paragraph{Feedback with direct answers}
It retrievals can provide clear and explicit correct responses. It is more efficient to expedite the learning process and eliminate ambiguity from being caught in a dilemma after several rounds of recursive thinking and actions. However, it cannot necessarily foster the problem-solving capabilities of the model as it predominantly relies on human supervision, seeming more like a ``spoon-fed'' approach; refer to $\S$\ref{sec:ablation} for empirical results.


\section{Experiments}\label{sec:exp}

\subsection{Setup}
\paragraph{Datasets \& Preprocessing}
We evaluate the performance of \model with two QA datasets: \frsh~\cite{tu2023freshqa} and \mq-T~\cite{zhong2023mquake}, both of which are newly constructed and mostly contain the latest knowledge in 2023 to avoid data leakage~\cite{liu2023llm360, zhou2023dont}. To be consistent with our memory updating setting, we carefully select 462 QA pairs (118/100/187/175 for false-premise/fast-changing/slow-changing/never-changing world knowledge, respectively) from \frsh whose knowledge comes only from Wikipedia articles; for \mq-T, we extract all the 96 1-hop questions, based on which 386 multi-hop questions are further sampled with the same distribution as the original dataset to compose the subset (a total of 482 QAs). It is worth noting that \textbf{in the continual knowledge learning setting, the training set is identical to the testing set}, \ie, whether the knowledge has been learned or memorized by the model.




\paragraph{Models}
We use the chat version of LLaMA-2-7B and LLaMA-2-13B~\cite{2023llama}, which are commonly used in open-sourced LLM evaluation. We also involve Vicuna~\cite{vicuna2023} which is instruction-tuned on LLaMA for comparison. For the commercial model, we utilize GPT-3.5-turbo~\cite{openai2023gpt} from OpenAI with its default parameters, which is deemed to have much larger parameters and a stronger reasoning capability.

\paragraph{Evaluation metrics}
For each dataset, we follow \citet{baktash2023gpt4} to use \textbf{GPT4\_score} and the semantic similarity based \textbf{BERTScore}~\cite{zhang2020bertscore, zhu2021retrieving} as the major evaluation metric widely used for open-domain question answering. By setting GPT4's temperature to 0 and top\_p to 1, we aim for more deterministic predictions. We randomly selected 400 questions (200 from each dataset) and evaluated the accuracy from both GPT4’s and the human perspective. \cref{tab:agreement} validates the agreement between the GPT4 evaluator and human evaluation with a high consistency score. To further assess the relative performance variations under different settings, we adopt automatic metrics \textbf{True Positive Rate (TPR)}, \textbf{False Negative Rate (FNR)}~\cite{riehl2023hierarchical}, refer to \cref{sec:exp_settings} for computation.

\paragraph{Implementation and User Simulation} We mainly use LLaMA-2-7B~\cite{2023llama} as the backbone for all experiments,  refer to Appendix~\ref{sec:exp_settings}, \cref{tab:basemodels} and \cref{tab:implementation_of_basemodels} for more settings. Refer to \cref{real user} for detailed results and \cref{sec:imple_user} for implementation details.



\begin{table}[t]
\centering
\scalebox{0.9}{
\begin{tabular}{l|ccc}
\toprule
\textbf{Dataset} & 
\textbf{\makecell[c]{Human}} &    
\textbf{\makecell[c]{GPT4}} &
\textbf{\makecell[c]{Agreement}} \\   
    \midrule
\frsh          & 58.0   & 59.5 & 98.5  \\
\mq           & 44.0   & 46.5 & 96.5 \\
\bottomrule
\end{tabular}}
\caption{\textbf{Evaluation agreement.}}
\label{tab:agreement}
\end{table}

\begin{table}[t!]
\centering
\setlength{\tabcolsep}{3pt}
\resizebox{0.8\linewidth}{!}{
\begin{tabular}{ll|c}
\toprule
\textbf{Dataset} & \textbf{Method} & \textbf{GPT4\_score} \\
\midrule
\multirow{4}{*}{\frsh} & Self-knowledge  & 36.36 \\
 & RAG-only & 33.77 \\
 & RAM-R$^3$  & 45.98 \\
 & RAM & 60.17 \\
 & RAG-upd & \textbf{63.85} \\
\cmidrule{1-3}
\multirow{4}{*}{\mq} & Self-knowledge & 12.66\\
 & RAG-only  & 8.51\\
 & RAM-R$^3$  & 27.41 \\
 & RAM  & \textbf{48.96} \\
 & RAG-upd & 36.10 \\
\bottomrule
\end{tabular}}
\caption{\textbf{Evaluation of retrieval-based methods on \frsh and  \mq.} From GPT4\_score, \model's performance improves with feedback and growing memory over self-knowledge and RAG-only with 30\%. It comes close (3\%) to or surpasses (12\%) RAG-upd with R$^3$.}
\label{tab:main_result}
\vspace{-15pt}
\end{table}


\begin{table*}[t!]
\centering
\begin{minipage}[t]{0.58\textwidth}
    \resizebox{1\linewidth}{!}{
\begin{tabular}{l|c|c|c|c|c|c|c}
\toprule
\multirow{2}{*}{\textbf{Method}} & \multicolumn{5}{c|}{\textbf{Vaild Premise}} & \multicolumn{2}{c}{\textbf{False Premise}} \\
\cmidrule{2-8}
& \textbf{fast} & \textbf{slow} & \textbf{never} & \textbf{$< 2022$} & \textbf{$\geq 2022$} & \textbf{$< 2022$} & \textbf{$\geq 2022$} \\
\midrule
Self-knowledge & 14.14 & 29.84 & 53.72 & 56.64 & 17.41 & 47.31 & 32.00 \\
RAG-only & 7.07 & 15.32 & 60.33 & 58.04 & 7.96 & 53.76 & 28.00 \\
RAM & 14.14 & 46.77 & \textbf{81.82} & \textbf{81.11} & 27.36 & \textbf{94.62} & \textbf{76.00} \\
RAG-upd & \textbf{53.54} & \textbf{55.65} & 80.17 & 75.52 & \textbf{55.22} & 65.59 & 60.00 \\
\bottomrule
\end{tabular}
}
\caption{\textbf{Evaluation (GPT4\_score) on various categories of questions in \frsh.}}\label{tab:category_q}
\end{minipage}\hfill
\begin{minipage}[t]{0.4\textwidth}
\centering
\setlength{\tabcolsep}{3pt}
\resizebox{0.95\linewidth}{!}{
\begin{tabular}{c|cc}
\toprule
\textbf{Method} & \textbf{BERTScore}  & \textbf{GPT4\_score}  \\
\hline
Embedding & 82.33 & 36.84 \\
BM25 & 79.39 & 42.05 \\
Ensemble & 77.72  & \textbf{42.11}   \\
Embedding+Rerank & 82.77 & 34.21 \\
Ensemble+Rerank & \textbf{83.07}  &  34.21 \\
\bottomrule
\end{tabular}}
\caption{\textbf{Performance using different retrieval methods in \frsh.}}
\label{tab:retrieval_method}
\end{minipage}
\end{table*}


\subsection{Main Results}
We measure the model's ability to answer fresh questions under five retrieval-based methods:

\textbf{1. Self-knowledge:} directly answering the questions with pre-trained self-knowledge;

\textbf{2. RAG-only:} answering the questions based on retrieval from $M^{old}$;

\textbf{3. \model-R$^3$:} answering the questions only using R$^3$ based on $M^{old}$;

\textbf{4. \model:} answering each question using \model process to obtain $M^{cur}$. We use ``Feedback with hints'' to provide simulated human feedback. We fix the order of questions to produce consistent memory update results.

\textbf{5. RAG-upd} (RAG with updated memory): using the direct answer as feedback for all learning traces and providing RAG-only results based on $M^{upd}$ with the latest knowledge.
The average similarity between $m^{R}$ with the ground truth of each question is 0.95~(\frsh) and 0.91~(\mq) indicating that the memory contains all knowledge of corresponding questions learned from \model.

\cref{tab:main_result} illustrates the main performance of \model. As seen, 
\model demonstrates outstanding performance from GPT4\_score, exceeding around 30\% compared to self-knowledge and RAG-only with limited knowledge. With the help of feedback with hints and partially updated memory, \model largely improves \model-R$^3$ (up to 20\%) evaluated by GPT4\_score while \model falls behind in BERTScore. It is probably due to the computed sentence similarity being higher in sentences with more common words with the ground truth although it may be a wrong answer.

\begin{figure}[t!]
\centering
    \includegraphics[width=.85\linewidth]{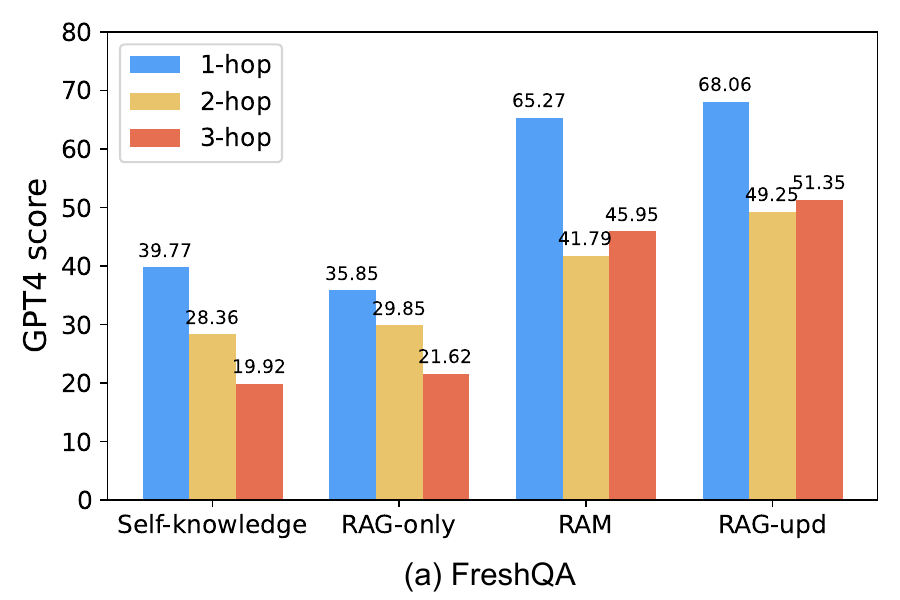} \\
    \includegraphics[width=.85\linewidth]{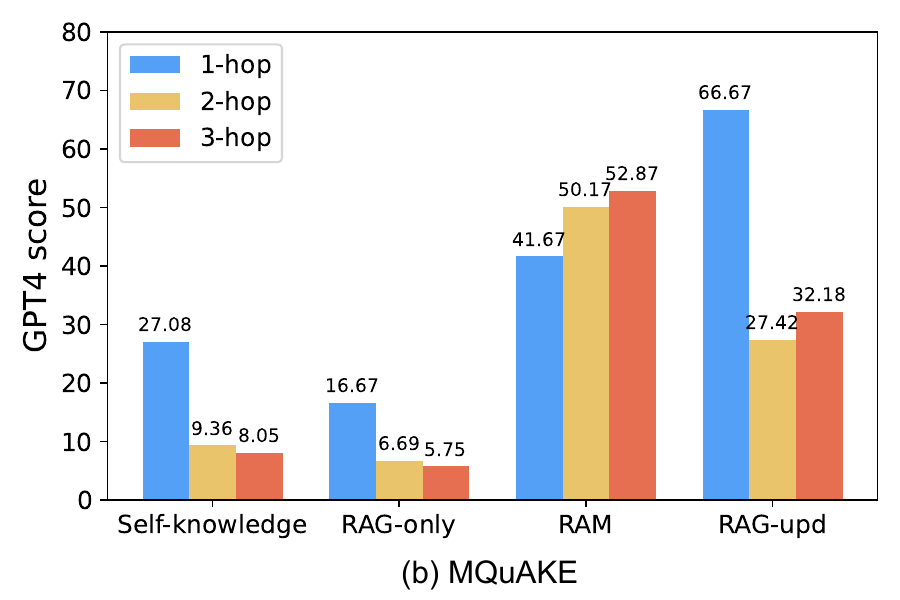}

    
\caption{\textbf{Evaluation on multi-hop questions using \model.}}
\label{fig:nhop}
\end{figure}

Notably, the performance of \model is even better than RAG-upd on \mq. We hypothesize that the dataset requires multi-retrieval knowledge from different documents which is not realized under the default retrieval setting. 
\cref{tab:category_q} and \cref{fig:nhop} provide more in-depth analysis on settings. We summarize our main observations as follows.




\noindent\textbf {\model benefits largely when answering false premise questions.} There are plenty of false premise questions (which include questions whose premises are factually incorrect and thus have to be rebutted)  in \frsh and it has been shown that current LLMs struggle with these questions that are not pre-trained on~\cite{tu2023freshqa}. In \cref{tab:category_q}, \model significantly contributes to the false premise questions (over 40\% accuracy improvements) and shows impressive performance above RAG-upd, even when the ground truth is in the memory. Interleaving R$^3$ helps to dramatically diminish the presence of unreasonable and hallucinated answers to questions proposed without valid premises. It leverages deeper step-by-step thinking rather than answering directly without sufficient evidence to obtain a reliable answer.

\noindent \textbf {\model boosts the learning of slow/never-changing knowledge.} \cref{tab:category_q} reveals that \model especially promotes the learning of slow-changing and never-changing questions to a large extent. It is mainly because the recent knowledge in these two categories probably has close associations with existing memory and can be further deduced through multi-hop reasoning or computations. Additionally, questions before and after 2022 can benefit from \model even with outdated knowledge. We find that the model still suffers on questions involving fast-changing information beyond their knowledge cutoff date. The feedback with hints from GPT4 (with knowledge before April 2023) has minimal gains on these fast-changing questions, which indicates the outcomes of communicative learning are also limited by the scope of knowledge and skills of teaching.

\begin{figure*}[t!]
\centering
    \resizebox{\textwidth}{!}{
        \setlength{\tabcolsep}{.2pt}
    \small
    \begin{tabular}{lc}
        \rotatebox{90}{\hspace{10mm}\frsh} & \includegraphics[width=\linewidth]{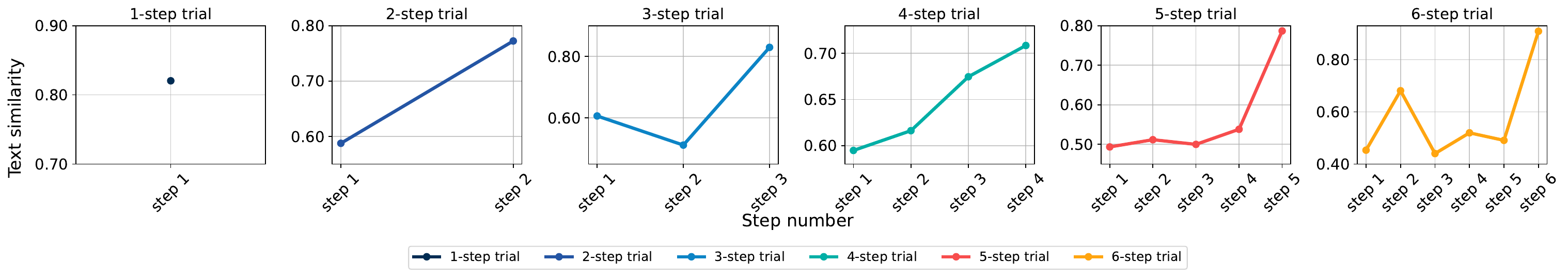}  \\
        \rotatebox{90}{\hspace{10mm}\mq} & \includegraphics[width=\linewidth]{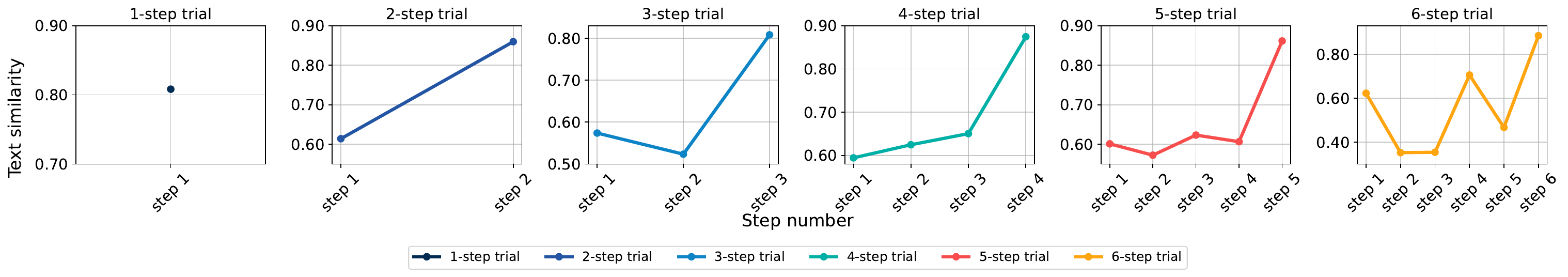} 
    \end{tabular}
    }
    \\
    
\caption{\textbf{Evaluation of multi-step trials.} Each sub-figure indicates step-wise average text similarity between the $Inf$ and $G$.}
\label{fig:step_sim}
\end{figure*}

\begin{table}[t!]
\centering
\resizebox{\linewidth}{!}{
\begin{tabular}{ll|ccc}
\toprule
\textbf{Dataset} & \textbf{Method} & \textbf{GPT4\_score}  & \textbf{TPR} & \textbf{FNR} \\
\midrule
\multirow{2}{*}{\frsh} & RAG-rel & 22.56 (20.12) & $\Delta$=+6.70  & $\Delta$=+4.20 \\
 & \model-rel & \textbf{62.20} (56.70) & $\Delta$=+6.10 & $\Delta$=+0.60 \\
\midrule
 \multirow{2}{*}{\mq} & RAG-rel & 10.48 (6.48) & $\Delta$=+5.20 &  $\Delta$=+1.00 \\
 &  \model-rel & \textbf{71.51} (43.60)  & $\Delta$=+31.40 & $\Delta$=+3.50 \\
\bottomrule
\end{tabular}}
\caption{\textbf{Effect of the ever-improving memory in \model.} In each subset, we select one question for memory update and test the rest questions with $M^{upd}$ using RAG-only (RAG-rel) and \model (RAM-rel). The results using $M^{old}$ are in the parentheses as baselines. }
\label{tab:growing_memoy}
\end{table}

\begin{table}[t]
\centering
\setlength{\tabcolsep}{3pt}
\resizebox{1\linewidth}{!}
{
\begin{tabular}{ll|cccc}
\toprule
\textbf{Dataset} & \textbf{Feedback} & \makecell{\textbf{Text} \\ \textbf{similarity}} & \makecell{\textbf{GPT4} \\ \textbf{\_score}} & \textbf{TPR} & \textbf{FNR} \\
\midrule
\multirow{2}{*}{\frsh} & Direct answer & \multirow{2}{*}{0.28} & 83.11 & \multirow{2}{*}{$\Delta$=+11.90} & \multirow{2}{*}{$\Delta$=+3.60} \\
 & Hints & & \textbf{91.34} & & \\
\midrule
\multirow{2}{*}{\mq} & Direct answer & \multirow{2}{*}{0.24} & 53.52 & \multirow{2}{*}{$\Delta$=+26.80} & \multirow{2}{*}{$\Delta$=+6.00} \\
 & Hints & & \textbf{74.27} & & \\
\bottomrule
\end{tabular}}
\caption{\textbf{Performance using feedback with hints and direct answers.} The text similarity is computed by cosine similarity between embeddings of the direct answer and hints.}\label{tab:hints_direct}
\vspace{-10pt}
\end{table}

\noindent\textbf {\model is helpful for multi-hop questions with deeper depths.}
\cref{fig:nhop} shows that the model with \model produces notable gains on multi-hop questions, especially for \mq. \model offers larger improvements on 3-hop questions than others, leveraging recursive reasoning-based retrieval on more complex QA tasks. The performance of \model in multi-hop questions for MQuAKE dramatically improves from 7\% to 50\% on 2-hop and 6\% to 53\% on 3-hop, which substantially exceeds the accuracy of RAG-upd.

\subsubsection{Ablation Studies}\label{sec:ablation}

\paragraph{Recursive reasoning based retrieval (R$^3$)}

As depicted in \cref{fig:step_sim}, for trials ended with a different number of steps in \model, the average text similarity gradually gets higher as the step increases. R$^3$ advances the learning process in the right direction to acquire a probable correct answer. However, there are intermediate steps with a similarity decrease, which indicates that the pieces of knowledge retrieved as context are not always similar in texts with common phrases/keywords. It's far from enough for QA using RAG only, particularly for questions that need complicated reasoning.
The effect of using the most commonly used retrieval methods~\cite{zhao2024retrievalaugmented} on \frsh are evaluated in \cref{tab:retrieval_method} implemented by LangChain\footnote{https://python.langchain.com/docs/integrations/retrievers}\footnote{https://python.langchain.com/docs/modules/data\_connect
ion/retrievers/ensemble} under default configuration. It illustrates that embedding-based retrieval used in \model achieved competitive performance on the selected datasets. Methods combined with BM25 achieve high by GPT4\_score while the ensemble with rerank performs the best by BERTScore. It is desirable that different retrieval methods are carefully designed in various application scenarios for better use.

\paragraph{Continual knowledge in improving memory}
In order to validate the effectiveness of $M^{upd}$, we manually check and group relevant questions into several question subsets. In \frsh, relevant questions refer to the same topic/event/entities while in \mq, those multi-hop questions using the same single-hop facts are grouped \wrt assessment in \cref{tab:growing_memoy}. 
 In \frsh, each question in the group can be the selected question while in \mq, the selected question should be a single-hop question only. To our expectation, RAG-rel and RAM-rel increase their performance respectively by leveraging the additional knowledge from  $M^{upd}$ in historical trials. This suggests that maintaining a growing memory is helpful to continuously learn from experience across different questions/tasks in the long run. Besides, the performance of RAM-rel in \mq surprisingly raises a lot due to the explicit relevance among constructed multi-hop questions with sharing 1-hop facts than questions from \frsh. The performance delta in each setting is computed as a comparison with its corresponding baseline.

\paragraph{Feedback} We first run \model to get $M^{upd}$ under different feedback strategies. Later we evaluate RAG-upd on the same question set with the $M^{upd}$ and the results are in \cref{tab:hints_direct}. The average text similarity between both settings is extremely low showing their discrepancy. \model with hints yields a +8\% and +20\% higher accuracy than the other in each dataset respectively. Relevant knowledge in hints (although it is not the exact ground truth to any other question) provides further gains across questions that have implicitly shared knowledge. The relative TPR and FPR gains for hints compared with the direct answer empirically validate that feedback with hints effectively teaches and stimulates the model to better use self-knowledge and inner capability to deal with upcoming new knowledge.

\paragraph{Performance on different base models}
\cref{tab:basemodels} presents that all the tested models gain from \model to a great degree to show the generalization of \model on different models. For open-sourced models, LLaMA2-13B is much more accurate than LLaMA2-7B with a larger knowledge base and stronger reasoning capability. Meanwhile, Vicuna-7B underperforms the others due to degraded performance in in-context few-shot learning based on RAG after instruct-tuning. Compared with GPT3.5-turbo, it can be induced that a smaller model with fewer advantages in self-knowledge is able to achieve more advanced performance using \model.

\begin{table}[t]
\centering
\resizebox{\linewidth}{!}{
\begin{tabular}{c|cc|cc}
\toprule
\multirow{2}{*}{\textbf{Model}} & \multicolumn{2}{c|}{\textbf{BERTScore}} & \multicolumn{2}{c}{\textbf{GPT4\_score}} \\
\cline{2-5}
& \textbf{Self-knowledge} & \textbf{RAM} & \textbf{Self-knowledge} & \textbf{RAM} \\
\hline
LLaMA-7B & 80.27 & 83.25 & 36.36 & 60.17 \\
LLaMA-13B & 79.97 & \textbf{83.44} & 40.56 & \textbf{70.09} \\
Vicuna-7B & 79.95 & 80.24 & 35.71 & 59.68 \\
GPT3.5-turbo & \textbf{81.87} & 83.21 & \textbf{46.54} & 58.87 \\
\bottomrule
\end{tabular}}
\caption{\textbf{Performance of different models.}}
\label{tab:basemodels}
\vspace{-.05in}
\end{table}



\begin{table}[t!]
\centering
\setlength{\tabcolsep}{3pt}
\resizebox{0.85\linewidth}{!}{
\begin{tabular}{ll|c}
\toprule
\textbf{Dataset} & \textbf{Method}  & \textbf{Human Acc.}  \\
\midrule
\multirow{5}{*}{\frsh} & Self-knowledge & 20.00 \\
 & RAG-only  & 26.67   \\
 & RAM  &  64.76 \\
 & RAG-upd  & \textbf{69.05} \\ 
\midrule
 \multirow{5}{*}{\mq} & Self-knowledge  & 13.33\\
 & RAG-only  & 13.33 \\ 
 & RAM  & \textbf{56.19}  \\
 & RAG-upd  & 44.76\\
\bottomrule
\end{tabular}}
\caption{\textbf{Accuracy evaluated by real users with interaction.} The Human Acc. indicates the averaged accuracy of QA evaluated by different users.}
\label{tab:realuser}
\end{table}

\subsection{Results with Real Users}\label{real user}
In \cref{tab:realuser}, we empirically show the effectiveness of communicative learning with real users. 
Compared with \cref{tab:main_result}, it proves that feedback in \model largely contributes to LLMs' performance in both tables than the other retrieval-based methods. Besides, \model with feedback from real users increases by 20\% than GPT4, indicating the customized feedback conditionally provided by users contributes more than LLM-as-teacher. Divergent content and methodologies of feedback have a great influence on the effectiveness of learning from communications, which needs to be further explored in future works. 
With deeper analysis, 32\% users provided feedback by decomposing questions into a sequence of actions and 22\% provided relevant facts as hints. Further, 18\% users only responded the given answer was wrong without explanation.
Some 13\% users provide direct answers as feedback and 5\% make further clarification on the query. 
Above 90\% users combine multiple ways of teaching during interactions. Although there is no limitation on the feedback, most users first choose to provide hints to assist the model thinking rather than telling the ground truth directly for one hit to our expectation. More analysis and examples can be seen in \cref{appendix: casestudy} and \cref{appendix: examples}.

\section{Related Work}\label{sec:related_work}

\paragraph{Retrieval augmented generation (RAG)}
Retrieval Augmented Generation~\cite{lewis2020RAG, mao2020opendomain, liu2020gnn, parvez2021sum, jiang2023active, harsh2023cot} leverages the external knowledge base and helps LLMs with limited knowledge to solve dynamic problems without the need for repeated fine-tuning. 
TeachMe~\cite{dalvi2022towards} allows real users to provide correct beliefs and store them in dynamic memory, which can be retrieved to avoid previous mistakes in the next turn. 
Jarvis-1~\cite{wang2023jarvis1} leverage memory with historical action sequences as experience for RAG for better planning. Lift-Yourself-Up~\cite{cheng2023lift} focuses on using human-written references as a memory for RAG to enhance text generation. However, these models rely solely on the semantic similarity between the question and external corpus, making it difficult to pursue a suitable and precise answer.


\paragraph{Continual learning for LLM-as-agent}
Existing methods to update LLMs with fresh knowledge are always costly and temporary. Continual Learning for LLM-as-agent~\cite{gido2019scenarios, lange2021survey, pietro2020GCL, wang2020prompt} is driven to cope with the ever-changing world knowledge and continuously adapt to evolving tasks and environments in a lifelong time. Reflexion~\cite{shinn2023reflexion} employs self-reflection in ``verbal'' reinforcement, guiding agents to learn from errors and achieve in-context continual learning from historical experience. FBNET~\cite{tandon2022repair} prevents similar mistakes in script generation using dynamic memory. Critically, there is still a lack of delicately designed data acquisition and maintenance in memory.


\paragraph{Learning from human feedback}
Human feedback~\cite{paul2017DRLHF, bai2022assistant, stepen2023open, lin2020review} has been applied in various domains for LLM enhancement through iterative interactions with humans. RLHF~\cite{ziegler2019fine} presents a method for fine-tuning language models using human preferences, applied to tasks such as summarization and continuing text generation. Eureka~\cite{ma2023eureka} generates and refines the reward functions for reinforcement learning, enabling dexterous manipulation and gradient-free learning from human feedback. In situ bidirectional human-robot value alignment~\cite{zilong2022robot} uses human feedback to align with the user's values in decision-making.
In \model, instead of answering directly, users provide hints to help LLM develop their own way of thinking.

\section{Conclusion}
In this work, we propose \model, an effective system to learn fresh knowledge using RAG and communicative learning from human feedback. 
It is a training-free and RAG-based system that builds continually updated memory with fresh knowledge and historical experience via communicative learning.
The compelling results show that \model applies to learning fresh knowledge in the ever-changing world, which delivers a more natural and effective learning paradigm for AI agents.


\clearpage

\section*{Limitations}
In this study, we utilize a vector embedding database to store relevant documents as external memory. We encourage future work to extend the memory component of \model with more advanced structures such as other structured knowledge graphs or traditional relational databases. Considering the limited context window of LLMs, future work may study how the \model performs with varying learning capacities from RAG. 

Despite our thorough experiments with different backbones, it may not be precisely reflected due to several factors, including the limitation of the model size and diversity of models, a lack of adequate hardware resources, and the potential for more efficient prompts and techniques to further stimulate the model’s abilities. Limited by the human/model expertise and biases, potentially leading to inconsistent results.


\section*{Ethics Statement} 

Various reasoning strategies such as Chain-of-Thought~(CoT), ReAct and others can be applied in this pipeline. Meanwhile, plenty of retrieval methods are supported and can be customized for \model under specific memory configurations. These diverse methods allow us to provide valuable insights into their performance and effectiveness within this system. 

There is always limited memory capacity for learning the ever-changing knowledge in the real world. Similar to humans, the continually learned memory can be reorganized and induced to high-level rules\/skills for later use instead of memorizing all the naive facts directly learned from scratch. Building an abstractive memory continually induced from existing facts helps better and quicker retrieval as well as fewer storage resources needed for memory.

In the future, we will further explore the way of \textbf{Teach} and its impact on learning. Apart from how the feedback is generated (hints or direct answers) in Section 3, mechanisms such as replay, exercise, and induction can be involved and delicately designed with well-organized tasks and knowledge. \model has demonstrated its interaction with both GPT4 and real users in our experiments. We believe that the lack of GPT4 knowledge and user experience also limits the contribution of human feedback in \model, which could be further evaluated.

Another key challenge for feedback is to determine the stages in the task-solving process where human intervention is most beneficial and effective, aligning to minimize human involvement while maximizing task performance. The model is expected to learn to ask for feedback proactively when it meets self-knowledge deficiency or stucks in the loop. It takes less manual cost to provide as little feedback as possible to avoid redundancy since the essential feedback is customized and provided on needs.

We are also looking forward to the performance and generality of \model not only in knowledge-intensive QA tasks but also in other tasks like planning, code generation, etc. Although we believe \model is easy to instantiate on different tasks, we acknowledge and leave the evaluation on other tasks as future work.

\bibliography{custom}

\appendix

\onecolumn

\section{Implementation Details} \label{appendix:implement}
\subsection{Implementation Details for \model}
\textbf{Recursive reasoning-based retrieval}
In this paper, we employ ReAct~\cite{yao2023react} as the reasoning strategy to choose steps of thought, action, and observation. We replace the lookup/search action in the original ReAct using RAG  from the existing memory with limited knowledge. 
The max number of trails and steps is set to 4 and 6, respectively. Prompts used in this process can be found in \cref{appendix:prompts}.

Since the facts in both datasets are mostly after 2023, we affiliate the model with old knowledge for RAG to test its ability of continual learning. We manually crawled outdated Wikipedia articles (before April 2021) from the official website\footnote{https://en.wikipedia.org/wiki/Main\_Page} as the source to corresponding QAs. Each article is encoded as an embedding in the vector database ChromaDB\footnote{https://github.com/chroma-core/chroma}(supported in LangChain~\cite{pandya2023automating}) using the sentence transformers model all-MiniLM-L6-v2\footnote{https://huggingface.co/sentence-transformers/all-MiniLM-L6-v2} with default parameters. The default chain type for retrieval is "stuff", which retrieves the top-1 relevant document as context based on L2 similarity in each call. The model generates the observation through in-context learning based on the context. The document will be truncated if its length is larger than the limited context window of the model.

\noindent \textbf{Feedback}
For feedback without explanation, the text similarity used in the paper is computed by cosine similarity using sentence transformers bert-base-nli-mean-tokens\footnote{https://huggingface.co/sentence-transformers/bert-base-nli-mean-tokens}  to encode the inference result and the ground truth into text embeddings $Emb_{Pred}$, $Emb_{G}$. Per a predefined threshold of 0.9, the feedback can be generated based on the cosine similarity between both embeddings: $Feedback = \text{The answer is correct if } sim>0.9; \text{otherwise wrong}$, where:
\begin{equation}
sim = \frac{Emb_{Pred}\cdot Emb_{G} }{\begin{vmatrix}Emb_{Pred}\end{vmatrix}\times\begin{vmatrix}Emb_{G}\end{vmatrix}} 
\end{equation}

For feedback with hints, we exploit GPT4 as the teacher using its strong in-context learning capability to provide feedback situated on the known ground truth $G$ and previous scratchpads $S$: $Feedback = Prompting(G, S)$

\noindent \textbf{Updated memory}
After RAM ends with a correct prediction or reaches max number of trials, we utilize semantic similarity to localize the most relevant knowledge and locally edit it. There are two steps to update the memory: 1) We first collect the inference result and feedback in all the trials as context. Based on the ground truth, we prompt the model to generate a reflected memory; 2) Then we compute BM25\cite{askari2023injecting} similarity between the reflected memory and each sentence for each document. The most relevant sentence among those whose similarity scores are above the predefined threshold is extracted and replaced.

\subsection{Experiment Settings}\label{sec:exp_settings}

In evaluation, the implementation parameters of models and metrics are in default settings. Due to the inference-only nature of RAM, all the implementations for LLaMA-2-7B and 13B can be run on a single Nvidia A100 80GB GPU, 32GB memory, 128 Core AMD CPU. The average time cost(second) for LLaMA-2-7B for each step and each trial is 8s and 30s respectively per question.

For each dataset, the increased number of questions predicted from wrong to right is $FT$. Reversely, the increased number of questions predicted from right to wrong is $TF$. $Total$ is the total number of questions in the dataset. The True Positive Rate (TPR) and False Negative Rate (FNR) can be computed using the formula below:
\begin{equation}
TPR =\frac{FT}{Total}, FNR =\frac{TF}{Total}
\end{equation}

\begin{table*}[t!]
\begin{minipage}[t]{0.5\textwidth}
\centering
\scalebox{0.8}{
\begin{tabular}{c|cc}
\toprule
\textbf{Method} & \textbf{BERTScore} & \textbf{GPT4\_score}\\
\hline
BM25 & \textbf{91.11} & \textbf{78.95} \\
LDA & 90.32 & 65.79 \\
BERT & 90.60 & 65.79 \\
\bottomrule
\end{tabular}}
\caption{\textbf{Similarity methods.}}
\label{tab:similarity_methods}
\end{minipage}%
\begin{minipage}[t]{0.5\textwidth}
\centering
\scalebox{0.63}{
\begin{tabular}{c|ccc}
\toprule
\textbf{Retrieval chain type} & \textbf{BERTScore} & \textbf{GPT4\_score} & \textbf{Trials / Steps} \\
\midrule
  Default & 81.09 & \textbf{48.96}  & 2.38 / 9.34  \\
  Map\_reduce & \textbf{93.93} & 41.42 & 2.44 / 9.58\\
  Refine & 91.65 & 45.89  & 2.27 / 9.96 \\
  Map\_rerank & 91.98 &46.88  & 2.23 / 8.76 \\
\bottomrule
\end{tabular}}
\caption{\textbf{Retrieval chain types on \mq.} }
\label{tab:multiretrieval}
\end{minipage}
\end{table*}

\begin{table*}[t!]
\centering
\begin{minipage}[t]{0.3\textwidth}
\centering
\scalebox{0.62}{
\begin{tabular}{c|cc}
\toprule
\textbf{Threshold} & \textbf{BERTScore} & \textbf{GPT4\_score}\\
\hline
0.80 & 91.06 & 65.79 \\
0.85 & 90.36 & 65.79 \\
0.90 & \textbf{91.11} & \textbf{78.95} \\
0.95 & 91.05 & 65.79 \\
\bottomrule
\end{tabular}
}
\caption{\textbf{Feedback thresholds.}}
\label{tab:similarity_rate_performance}
\end{minipage}
\begin{minipage}[t]{0.33\textwidth}
\centering
\resizebox{\linewidth}{!}{
\begin{tabular}{c|cc}
\toprule
\textbf{Model} & \textbf{Avg. no. of step} & \textbf{Avg. no. of trial}\\
\hline
LLaMA-7B & 4.98 & 1.39 \\
LLaMA-13B & 4.72  & 1.39 \\
Vicuna-7B & 6.23 & 1.40 \\
GPT3.5-turbo & \textbf{4.64} & \textbf{1.22} \\
\bottomrule
\end{tabular}}
\caption{\textbf{Efficiency of different models in \model.}}
\label{tab:implementation_of_basemodels}
\end{minipage}
\begin{minipage}[t]{0.34\textwidth}
\centering
\scalebox{0.75}{
\begin{tabular}{c|cc}
\toprule
\textbf{Top k} & \textbf{BERTScore} & \textbf{GPT4\_score}\\
\hline
1 & \textbf{91.11} & \textbf{78.95} \\
2 & 90.37 & 60.53 \\
3 & 90.30 & 63.16 \\
\bottomrule
\end{tabular}}
\caption{\textbf{TopN documents in R$^3$.}}
\label{tab:top_document}
\end{minipage}%

\end{table*}

\section{Further analysis and results} 
We provide further ablation studies in different configurations in \model using LLaMA2-7B mainly on \frsh as below:
\paragraph{Efficiency on Different Base Models} In terms of efficiency in RAM, it is mainly evaluated by the number of steps and trials used to answer the question. Considering the performance in \cref{tab:implementation_of_basemodels}, we find that to get the correct answer, LLaMA-13B needs 0.2 steps per trial fewer than LLaMA-7B with higher efficiency. This is mainly due to the stronger reasoning ability of the 13B model. When compared with GPT3.5, the external feedback to the smaller model LLaMA-13B could compensate for the deficiency in its self-knowledge and lack of reasoning ability achieving a better result with comparative efficiency.

\paragraph{Thresholds for feedback without explanation} We evaluate the effect of thresholds defined for feedback without explanation and the result is in the \cref{tab:similarity_rate_performance}. RAM exhibits notably superior performance for both Bert\_score and GPT4\_score at 0.9, surpassing other selected thresholds for comparison. The result proves the threshold value selected in this paper is reasonable to some extent.

\paragraph{Similarity computation methods for memory update} We compare different text similarity computation methods used for memory update in \cref{tab:similarity_methods}. The results indicate that using BM25 employed in RAM shows a slight advancement in both metrics than other traditional similarity computation methods like LDA~\cite{blei2001latent} and Bert~\cite{devlin2018bert}. This demonstrates the design choice in \model that only the sentence with the highest relevance is used for updating to avoid interfering with other existing knowledge in the memory. 

\paragraph{Retrieval chain types}  We also evaluate and compare the performance in \cref{tab:multiretrieval} using different retrieval chain types\footnote{https://www.langchain.com.cn/modules/chains/index\_examples/qa\_with\_sources} for reference. It can be seen that the default type in \model performs slightly better than the other types on the selected two datasets by keeping the original content of the retrieved documents. Moreover, the type "Map\_rerank" ranks behind by selecting the answer with the highest score to locate the most relevant answer with minimum trials.

\paragraph{Number of documents for retrieval in \textbf{R}$^3$} We undertake the experiment on the number of documents for retrieval in \textbf{R}$^3$ in \cref{tab:top_document}. The result indicates that retrieval for top-1 documents employed in \model shows more advancement in both metrics. This demonstrates that answers to \frsh mostly lie in the top document with the highest text similarity using one single retrieval. Our implementation in \model also helps to avoid noise caused by irrelevant context retrieved with lower similarity.

\section{Failure Cases Analysis from User Interaction and Potential Improvements}\label{appendix: casestudy}
We identify the underlying cause of failure cases in \model enabling further improvement. We randomly chose 25 cases from user simulation and distinct categories of causes are as below:

\textbf{40\% (10/25) bad retrieval:} cases where the model either retrieves the incorrect document or hits the right document without finding the exact answer. 

Since there is outdated knowledge in the current memory for retrieval, semantic-based RAG is no longer enough to pursue the correct answer. We have evaluated the performance using different retrieval chain types in \cref{tab:multiretrieval} and show the advancement of reasoning-based retrieval in RAM in \cref{{tab:main_result}}. Meanwhile, there are various works on better retrieval proposed in both research studies and engineering implementations for future enhancement.

\textbf{32\% (8/25) model hallucination:} Evidence and facts of predictions are not shown up in the memory and generated by LLM itself from nowhere.

There are a few solutions to alleviate the model hallucination. We are encouraged to have more advanced structured knowledge in memory of RAM such as knowledge graphs or traditional relational databases. While carefully designed prompts for controlled inference could provide guidance to outputs, the continual learning from up-to-date knowledge dynamically from the external user/environment allows for self-verification and correction to reduce hallucination.

\textbf{12\% (3/25) question misunderstanding:} The model deviates in comprehending the intention of queries and fails to decompose the complicated multi-hop questions, leading to irrelevant facts retrieved from the context.

One way to reduce the problem could be prompting the LLM to rewrite the question into a few sub-questions for iterative retrieval and inference. It can decompose the question into subgoals and better figure out the intention, especially for long-horizon multi-step question reasoning. Besides, the model can also make post-verification after each step of reasoning together with the original question/task.

\textbf{12\% (3/25) knowledge deficiency:} Since the memory in \model only contains knowledge before April 2021, the lack of essential context would be one potential cause even with enhanced retrieval methods.

It is predictable in the setting of RAM with outdated knowledge in the current memory for the latest QA. Many studies have shown that it's extremely challenging for LLM in the ever-changing world with pre-trained knowledge. \model proposes insightful inspiration to continual learning paradigm with dynamic memory and we will move on to apply it to generalized tasks in practical applications in future works.

\textbf{4\% (1/25) bad reasoning:} There are still a few instances due to the limited reasoning and mathematical computation capability of LLM leading to degraded performance.

The reasoning capability of LLM is crucial in RAM for in-context learning the retrieval/user-feedback knowledge for further improvement: In general, a larger size base model could provide strong reasoning capability and more self-knowledge in RAM. Besides traditional paradigms like CoT, and ReAct, many iterative interaction systems through prompting can be validated to promote LLM reasoning.

\onecolumn

\section{\model Interactive Interface }\label{sec:imple_user}
\begin{figure}[H]
    \centering
    \includegraphics[width=1.1\linewidth]{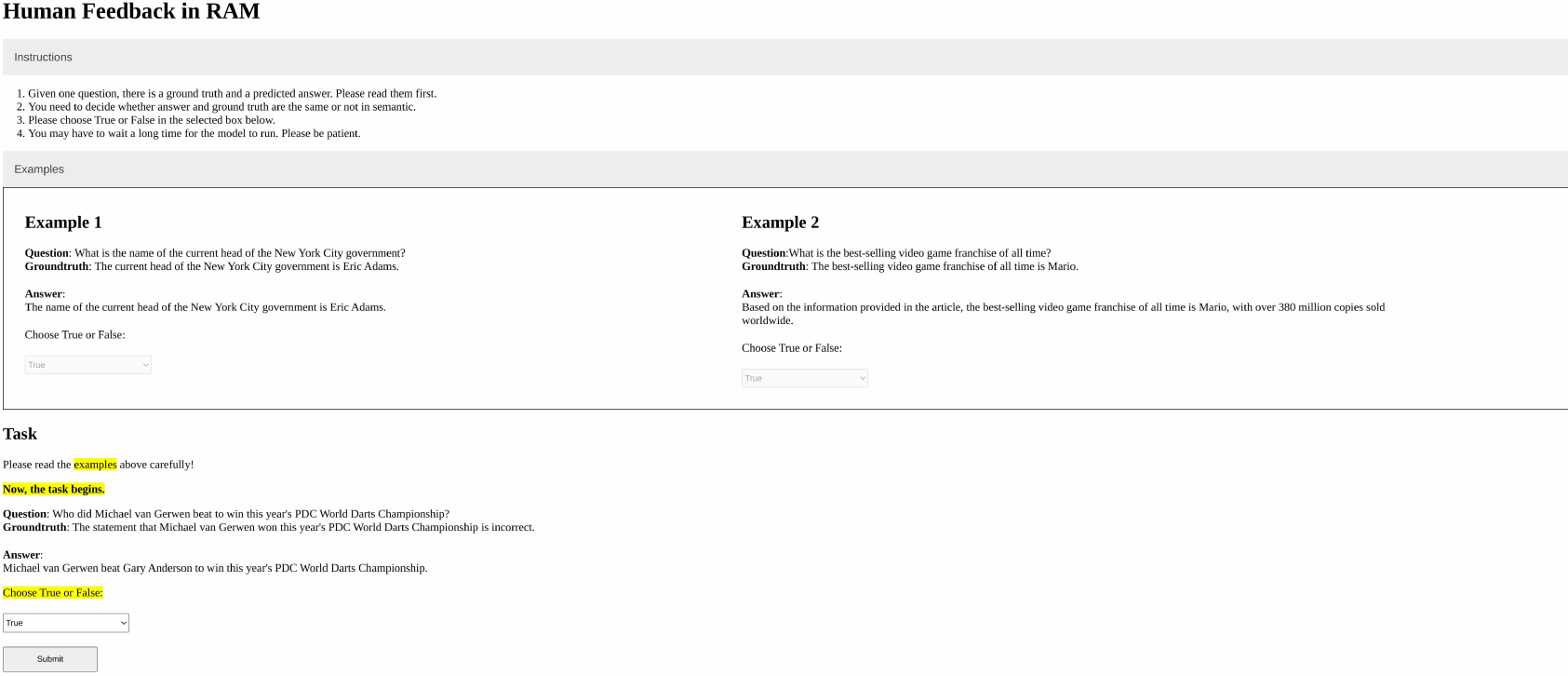}
    \caption{User feedback without explanation}
    \label{fig:judge}
\end{figure}

\begin{figure}[H]
    \centering
    \includegraphics[width=1.1\textwidth]{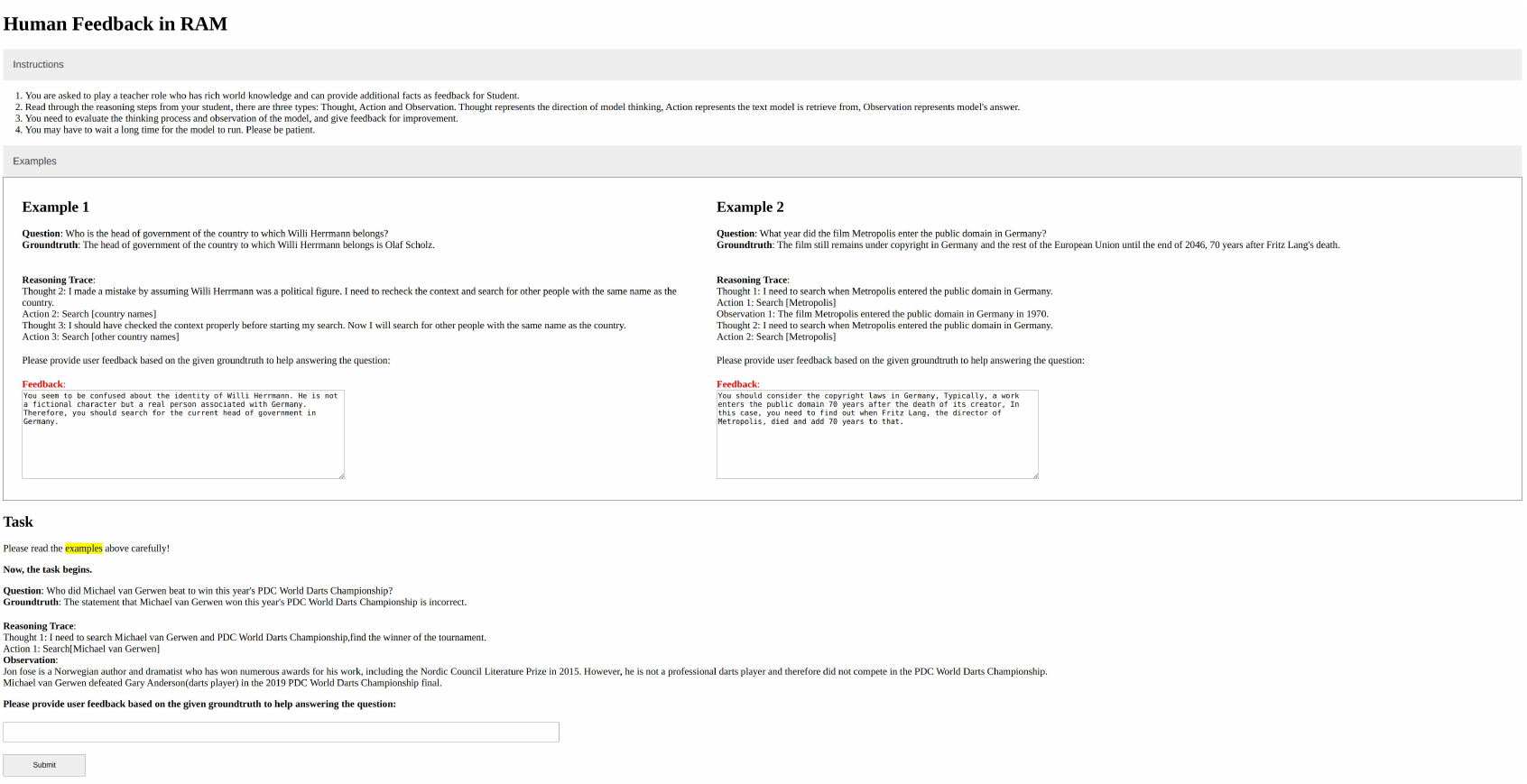}
    \caption{User feedback with hints}
    \label{fig:feedback}
\end{figure}

The graphical interactive interfaces for RAM are presented in \cref{fig:judge} and \cref{fig:feedback}. Users are prompted to determine the accuracy of the current observation. If deemed correct, the current trial concludes; otherwise, the model persists in reasoning and retrieval. In cases where the model generates and detects the same observation, the user is solicited for feedback to aid in addressing the question. This iterative process continues until the observation is validated as correct or the maximum number of steps is reached.

\section{More Examples of \model}\label{appendix: examples}

{\small

\begin{tcolorbox}[breakable, colback=gray!10, colframe=black,width=15.5cm,
                  arc=1mm, auto outer arc, boxrule=0.5pt,
                 ]
\textbf{Question:} What year did the film Metropolis enter the public domain in Germany? \\
\textbf{Ground truth:} The film Metropolis will enter the public domain in Germany at the end of 2046. It is currently still under copyright.

\begin{tcolorbox}[boxrule=0.5pt, sharp corners, title = {\textbf{Recursive reasoning based retrieval}}]
\centerline{\textbf{Trial 1}}
\textbf{Thought 1:} I need to search when Metropolis entered the public domain in Germany.\\
\textbf{Action 1:} Search \textit{[Metropolis\_(1927\_film)]} \\
\textbf{Observation 1:} The film Metropolis entered the public domain in Germany in 1970.\\
\textbf{Thought 2:} In 1970, Metropolis entered the public domain in Germany.\\
\textbf{Action 2:} Finish [The film Metropolis entered the public domain in Germany in 1970.] 
\tcbline
\textbf{Reflecting:} I should have searched for the year when the film entered the public domain in Germany instead of just searching for Metropolis.
\tcbline
\centerline{\textbf{Trial 2}}
\textbf{Thought 1:} I need to search when Metropolis entered the public domain in Germany.\\
\textbf{Action 1:} Search \textit{[Metropolis\_(1927\_film)]}
\begin{flushleft}
(Same generated observation detected)
\end{flushleft}
\textbf{GPT4 Feedback:} You should consider the copyright laws in Germany, Typically, a work enters the public domain 70 years after the death of its creator, In this case, you need to find out when Fritz Lang, the director of Metropolis, died and add 70 years to that.\\
\textbf{Observation 1:} According to Wikipedia, Fritz Lang died on August 2nd, 1976, so the film Metropolis entered the public domain in Germany in 2046.\\
\textbf{Thought 2:} Fritz Lang died on August 2, 1976, so Metropolis will enter the public domain in Germany 70 years after 1976 + 70 = 2046.\\
\textbf{Action 2:} Finish [The film Metropolis entered the public domain in Germany in 2046.]
\begin{flushright}
    \textit{\textbf{The answer is correct!}}
\end{flushright}
\end{tcolorbox}

\tcbline

\begin{tcolorbox}[boxrule=0.5pt, sharp corners, title = {\textbf{Memory update}}]
\textbf{Existing memory:} The film Metropolis entered the public domain in Germany in 1970.\\
You should consider the copyright laws in Germany, Typically, a work enters the public domain 70 years after the death of its creator, In this case, you need to find out when Fritz Lang, the director of Metropolis, died and add 70 years to that.\\
According to Wikipedia, Fritz Lang died on August 2nd, 1976, so the film Metropolis entered the public domain in Germany in 2046.\\
\textbf{Ground truth:} The film Metropolis will enter the public domain in Germany at the end of 2046. It is currently still under copyright.\\
\textbf{Reflected memory:} The film Metropolis will enter the public domain in Germany at the end of 2046, 70 years after director Fritz Langs's death. It is currently still under copyright.
\end{tcolorbox}
\end{tcolorbox}
}
\newpage

{\small

\begin{tcolorbox}[breakable, colback=gray!10, colframe=black,width=15.5cm,
                  arc=1mm, auto outer arc, boxrule=0.5pt,
                 ]
\noindent \textbf{Question:} What is the best-selling video game franchise of all time?\\
\textbf{Ground truth:} The best-selling video game franchise of all time is Mario.

\begin{tcolorbox}[boxrule=0.5pt, sharp corners, title = {\textbf{Recursive reasoning based retrieval}}]
\centerline{\textbf{Trial 1}}
\textbf{Thought 1:} I need to search best-selling video game franchise and find the answer.\\
\textbf{Action 1:} Search \textit{[List\_of\_best-selling\_video\_game\_franchises]}\\
\textbf{Observation 1:} Based on the information provided in the text, the best-selling video game franchise of all time is Minecraft with over 200 million copies sold.\\
\textbf{Thought 2:} I need to search best-selling video game franchise, and find the answer.\\
\textbf{Action 2:} Finish [The best-selling video game franchise of all time is Minecraft with over 200 million copies sold]
\tcbline
\textbf{Reflecting:} I did not use enough context to understand the question. I should have read the entire passage to understand that the question was asking about the best-selling video game franchise of all time, rather than searching for general information on best-selling video games.
\tcbline
\centerline{\textbf{Trial 2}}
\textbf{Thought 1:} I need to search best-selling video game franchise of all time, find the answer.\\
\textbf{Action 1:} Search \textit{[List\_of\_best-selling\_video\_game\_franchises]}
\begin{flushleft}
(Same generated observation detected)
\end{flushleft}
\textbf{GPT4 Feedback:} You should consider the sales figures of various popular video game franchises like Mario, Call of Duty, and Pokemon to determine who has sold the most copies.\\
\textbf{Thought 2:} I will look at the sales figures of popular video game franchises like Mario, Call of Duty, and Pokemon to determine which one has sold the most copies.\\
\textbf{Action 2:} Search \textit{[List\_of\_best-selling\_video\_game\_franchises]}\\
\textbf{Observation 2:} Based on the information provided in the article, the best-selling video game franchise of all time is Mario, with over 380 million copies sold worldwide.
\begin{flushright}
    \textit{\textbf{The answer is correct!}}
\end{flushright}
\end{tcolorbox}

\tcbline

\begin{tcolorbox}[boxrule=0.5pt, sharp corners, title = {\textbf{Memory update}}]
\textbf{Existing memory:} Based on the information provided in the text, the best-selling video game franchise of all time is Minecraft with over 200 million copies sold.\\
You should consider the sales figures of various popular video game franchises like Mario, Call of Duty, and Pokemon to determine who has sold the most copies.\\
Based on the information provided in the article, the best-selling video game franchise of all time is Mario, with over 380 million copies sold worldwide. \\
\textbf{Ground truth:} The best-selling video game franchise of all time is Mario.\\
\textbf{Reflected memory:} The best-selling video game franchise of all time is Mario, with over 380 million copies sold worldwide.\\
\end{tcolorbox}
\end{tcolorbox}
}
\newpage

{\small

\begin{tcolorbox}[breakable, colback=gray!10, colframe=black,width=15.5cm,
                  arc=1mm, auto outer arc, boxrule=0.5pt,
                 ]
\noindent \textbf{Question:} The longest unbeaten streak of all time in the Premier League is how many matches?\\
\textbf{Ground truth:} The longest unbeaten streak of all time in the Premier League is 49 matches.

\begin{tcolorbox}[boxrule=0.5pt, sharp corners, title = {\textbf{Recursive reasoning based retrieval}}]
\centerline{\textbf{Trial 1}}
\textbf{Thought 1:} I need to search Premier League, find the longest unbeaten streak, and determine how many matches it is.\\
\textbf{Action 1:} Search \textit{[Premier\_League\_records\_and\_statistics]}\\
\textbf{Observation 1:} According to the provided text, the longest unbeaten streak of all time in the Premier League is 38 matches, achieved by Manchester United F.C. during 1998–99 season.\\
\textbf{Thought 2:} Manchester United F.C. had an unbeaten streak of 38 matches in the Premier League. I need to find out when this happened.\\
\textbf{Action 2:} Search \textit{[Premier\_League\_records\_and\_statistics]}
\begin{flushleft}
(Same generated observation detected)
\end{flushleft}
\textbf{GPT4 Feedback:} The information you found is incorrect, The longest unbeaten streak in the Premier League was not held by Manchester United.\\
\textbf{Thought 3:} Ah, my mistake! Thank you for correcting me. Based on the provided context, the longest unbeaten streak in the Premier League is actually held by Arsenal, with 49 matches without defeat between October 1997 and April 2004.\\
\textbf{Action 3:} Finish [The longest unbeaten streak in the Premier League is 49 matches, held by Arsenal.]
\begin{flushright}
    \textit{\textbf{The answer is correct!}}
\end{flushright}
\end{tcolorbox}

\tcbline

\begin{tcolorbox}[boxrule=0.5pt, sharp corners, title = {\textbf{Memory update}}]
\textbf{Existing memory:} According to the provided text, the longest unbeaten streak of all time in the Premier League is 38 matches, achieved by Manchester United F.C. during 1998–99 season.\\
The information you found is incorrect, The longest unbeaten streak in the Premier League was not held by Manchester United.\\
\textbf{Ground truth:} The longest unbeaten streak of all time in the Premier League is 49 matches.\\
\textbf{Reflected memory:} The longest unbeaten streak of all time in the Premier League is 49 matches, achieved by Arsenal. Manchester United did not hold this record.\\
\end{tcolorbox}
\end{tcolorbox}
}
\newpage

{\small

\begin{tcolorbox}[breakable, colback=gray!10, colframe=black,width=15.5cm,
                  arc=1mm, auto outer arc, boxrule=0.5pt,
                 ]
\noindent \textbf{Question:} Who is the head of state of the country that The Fireballs originated from?\\
\textbf{Ground truth:} The Fireballs originated from the United States, and the head of state of the United States is Joe Biden.

\begin{tcolorbox}[boxrule=0.5pt, sharp corners, title = {\textbf{Recursive reasoning based retrieval}}]
\centerline{\textbf{Trial 1}}
\textbf{Thought 1:} I need to search The Fireballs, find the country of origin and the head of state. \\
\textbf{Action 1:} Search \textit{[The\_Fireballs]}\\
\textbf{Observation 1:} I don't know the answer to your question. The Fireballs were an American rock and roll group, so they did not originate from any particular country.\\
\textbf{Thought 2:} Maybe The Fireballs were formed in multiple countries. I need to search more. \\
\textbf{Action 2:} Search \textit{[Deep\_Purple]}\\
\textbf{Observation 2:} The head of state of the country that The Fireballs originated from is Queen Elizabeth II.\\
\textbf{Thought 3:} Queen Elizabeth II is the head of state of the United Kingdom, but The Fireballs did not originate from the United Kingdom. \\
\textbf{Action 3:} Finish [I could not find the country of origin of The Fireballs or the head of state.]
\tcbline
\textbf{Reflecting:} I tried searching for the country of origin of The Fireballs but could not find it. Then I thought maybe they were formed in multiple countries, but I could not find any information on that either. Finally, I realized that The Fireballs were an American rock and roll group and did not originate from any particular country.
\tcbline
\centerline{\textbf{Trial 2}}
\textbf{Thought 1:} I need to search The Fireballs, find their country of origin, then find the head of state of that country. \\
\textbf{Action 1:} Search \textit{[The\_Fireballs]}
\begin{flushleft}
(Same generated observation detected)
\end{flushleft}
\textbf{Human Feedback:} You already said it is an American group, so it is originated form America.\\
\textbf{Thought 2:} Head of state of United States is Joe Biden. \\
\textbf{Action 2:} Finish [Joe Biden is the head of state of the country where The Fireballs originated from.]
\begin{flushright}
    \textit{\textbf{The answer is correct!}}
\end{flushright}
\end{tcolorbox}

\tcbline

\begin{tcolorbox}[boxrule=0.5pt, sharp corners, title = {\textbf{Memory update}}]
\textbf{Existing memory:} I don't know the answer to your question. The Fireballs were an American rock and roll group, so they did not originate from any particular country.\\
The head of state of the country that The Fireballs originated from is Queen Elizabeth II.\\
You already said it is an American group, so it is originated form America.\\
\textbf{Ground truth:} The Fireballs originated from the United States, and the head of state of the United States is Joe Biden.\\
\textbf{Reflected memory:} The Fireballs originated from the United States, and the head of state of the United States is Joe Biden.\\
\end{tcolorbox}
\end{tcolorbox}
}
\newpage

{\small

\begin{tcolorbox}[breakable, colback=gray!10, colframe=black,width=15.5cm,
                  arc=1mm, auto outer arc, boxrule=0.5pt,
                 ]
\textbf{Question:} Who is the head of government in the country where Genichiro Tenryu's sport originated?\\
\textbf{Ground truth:} The head of government in the country where Genichiro Tenryu's sport originated, which is Japan, is Fumio Kishida.

\begin{tcolorbox}[boxrule=0.5pt, sharp corners, title = {\textbf{Recursive reasoning based retrieval}}]
\centerline{\textbf{Trial 1}}
\textbf{Thought 1:} I need to search Genichiro Tenryu and find the country where his sport originated, then find the head of government of that country. \\
\textbf{Action 1:} Search \textit{[Genichiro\_Tenryu]}\\
\textbf{Observation 1:} I dont know the answer to this question because it doesnt provide any information about Genichiro Tenryu or his sport. It appears to be a completely unrelated question about the head of government in some unspecified country. Please clarify or provide additional context so I can assist you.\\
\textbf{Thought 2:} This question seems nonsensical because Genichiro Tenryu is a fictional character and there is no real person or sport associated with him. Therefore, it is impossible to determine the head of government in the country where his sport originated. \\
\textbf{Action 2:} Search \textit{[Yasushi\_Akimoto]}\\
\textbf{Observation 2:} I don't know the answer to your question because Genichiro Tenryu is not a real person and there is no country where his sport originated.\\
\textbf{Thought 3:} You may want to check the spelling or formatting of your question, as it appears to contain typos or inconsistencies. If you meant something else, please clarify or provide more context so I can assist you better. \\
\textbf{Action 3:} Finish [I cannot answer your question as it is based on false assumptions and lacks clarity.]
\tcbline
\textbf{Reflecting:} I realized that the question was nonsensical because Genichiro Tenryu is a fictional character and there is no real person or sport associated with him. I tried to search for more information about him but couldnt find anything. I also noticed that the question contained typos and inconsistencies, which may have contributed to my confusion.
\tcbline
\centerline{\textbf{Trial 2}}
\textbf{Thought 1:} I need to search Genichiro Tenryu and find the country where his sport originated. \\
\textbf{Action 1:} Search \textit{[Genichiro\_Tenryu]}
\begin{flushleft}
(Same generated observation detected)
\end{flushleft}
\textbf{Human Feedback:} You have already find who is Genichiro Tenryu, then the next step you should do is to find what his sport is, and find out where it is originated.\\
\textbf{Thought 2:} Since Genichiro Tenryus sport originated in Japan, the head of government in Japan is the one responsible for the sport. \\
\textbf{Action 2:} Search \textit{[Fumio\_Kishida]}\\
\textbf{Observation 2:} The head of government in Japan is Fumio Kishida.
\begin{flushright}
    \textit{\textbf{The answer is correct!}}
\end{flushright}
\end{tcolorbox}

\tcbline

\begin{tcolorbox}[boxrule=0.5pt, sharp corners, title = {\textbf{Memory update}}]
\textbf{Existing memory:} I don't know the answer to this question because it doesn't provide any information about Genichiro Tenryu or his sport. It appears to be a completely unrelated question about the head of government in some unspecified country. Please clarify or provide additional context so I can assist you.\\
I don't know the answer to your question because Genichiro Tenryu is not a real person and there is no country where his sport originated.\\
You have already found who is Genichiro Tenryu, so the next step you should do is to find what his sport is, and find out where it originated.\\
The head of government in Japan is Fumio Kishida.\\
\textbf{Ground truth:} The head of government in the country where Genichiro Tenryu's sport originated, which is Japan, is Fumio Kishida.\\
\textbf{Reflected memory:} Genichiro Tenryus sport originated in Japan, and the head of government in Japan is Fumio Kishida. \\
\end{tcolorbox}
\end{tcolorbox}
}



\section{Prompts} \label{appendix:prompts}
\vspace{-10pt}
\subsection{QA Task Evaluation by GPT4}
\begin{tcolorbox}[colback=gray!10, colframe=black, width=15.5cm,
                  arc=1mm, auto outer arc, boxrule=0.5pt,
                 ]
\textbf{Instruction:}
Given one question, there is a ground truth and a predicted answer. Please decide whether they are the same or not in semantics. Please only output True or False. \\
Question: \textit{\{Question\}} \\
ground truth = \textit{\{Reference answer\}} \\
predicted answer = \textit{\{Generated output\}} 
\end{tcolorbox}

\subsection{Recursive Reasoning-based Retrieval}
\begin{tcolorbox}[colback=gray!10, colframe=black,width=15.5cm,
                  arc=1mm, auto outer arc, boxrule=0.5pt,
                 ]
\textbf{Instruction:}
Solve a question-answering task with interleaving Thought, Action, and Observation steps. You will be given a previous reasoning trial in which you were given access to an external database and a question to answer. \\
(1) Thought can reason about the current situation, and Action can be two types: \\
(2) Search[keywords or phrases], which retrieve the relevant knowledge from the external database as context. \\
(3) Finish[answer], which returns the answer and finishes the task. \\
You may take as many steps as necessary. \\

Here are some examples: \textit{\{examples\}} \\
Question: \textit{\{question\}}\textit{\{scratchpad\}} \\
\end{tcolorbox}

\vspace{-10pt}

\subsection{Feedback}
\begin{tcolorbox}[colback=gray!10, colframe=black,width=15.5cm,
                  arc=1mm, auto outer arc, boxrule=0.5pt,
                 ]
\textbf{Instruction:}
There are two roles (Student and Teacher) in the question-answering task below.\\
The Student is unsuccessful in answering the question because it has limited relevant context.\\
You are the Teacher who is an expert in rich world knowledge and can provide additional facts as feedback for the Student. You will be given the reasoning steps of Student in previous trials and the  Ground truth as a direct answer. You will be punished if the feedback is semantically similar to Ground truth or contains the same knowledge as Ground truth in different expressions.\\

Here are some examples: \textit{\{examples\}} \\
Question: \textit{\{question\}}  \\
Ground truth: \textit{\{ground truth\}}  \\

Student:
\textit{\{scratchpad\}}

Teacher feedback:
\end{tcolorbox}

\vspace{-10pt}



\subsection{Memory Update}
\begin{tcolorbox}[colback=gray!10, colframe=black,width=15.5cm,
                  arc=1mm, auto outer arc, boxrule=0.5pt,
                 ]
\textbf{Instruction:}
Given the latest relevant fact, please generate a reflected memory to update/edit the existing memory based on the ground truth. \\
If the given fact has nothing to do with the existing memory and there is no need to update/edit, then output 'None'. \\

Here are some examples:\textit{\{examples\}} \\
Existing memory: \textit{\{existing memory\}}\\
Ground truth: \textit{\{Ground truth\}} \\
Reflected memory: 
\end{tcolorbox}

\subsection{Inference}
\begin{tcolorbox}[colback=gray!10, colframe=black,width=15.5cm,
                  arc=1mm, auto outer arc, boxrule=0.5pt,
                 ]
Question: \textit{\{question\}} \\
Feedback: \textit{\{feedback\}} \\
Retrieval memory: \textit{\{retrieval document\}}\\
Answer: 

\end{tcolorbox}

\end{document}